\documentclass[10pt,twocolumn,letterpaper]{article}

\usepackage{cvpr}              

\usepackage{graphicx}
\usepackage[accsupp]{axessibility}  
\usepackage{amsmath}
\usepackage{amssymb}
\usepackage{booktabs}
\usepackage{multirow}
\usepackage[table]{xcolor}
\definecolor{Red}{RGB}{230, 57, 70}

\usepackage{pifont}
\usepackage[pagebackref,breaklinks,colorlinks]{hyperref}

\usepackage[capitalize]{cleveref}
\crefname{section}{Sec.}{Secs.}
\Crefname{section}{Section}{Sections}
\Crefname{table}{Table}{Tables}
\crefname{table}{Tab.}{Tabs.}

\def\cvprPaperID{3626} 
\def\confName{CVPR}
\def\confYear{2023}

\begin{document}


\title{  Semantic Human Parsing via Scalable Semantic Transfer \\ over Multiple Label Domains }

\author{Jie Yang$^{1}$~,~Chaoqun Wang$^{1}$~,~Zhen Li$^{1}$,~Junle Wang$^{2}$,~Ruimao Zhang$^{1}$\thanks{Corresponding author.} \\ 
$^1$The Chinese University of Hong Kong, Shenzhen,
$^2$Tencent \\
\texttt{\small{\{jieyang5@link,chaoqunwang@link,lizhen@,zhangruimao@\}cuhk.edu.cn}}\\
\texttt{\small{\{wangjunle@\}gmail.com}}\\
}

\maketitle


\begin{abstract}

This paper presents Scalable Semantic Transfer (SST), a novel training paradigm, to explore how to leverage the mutual benefits of the data from different label domains (\textit{i.e.} various levels of label granularity) to train a powerful human parsing network.
In practice, two common application scenarios are addressed, termed \textbf{universal parsing} and \textbf{dedicated parsing}, 
where the former aims to learn homogeneous human representations from multiple label domains and switch predictions by only using different segmentation heads, 
and the latter aims to learn a specific domain prediction while distilling the semantic knowledge from other domains.  
The proposed SST has the following appealing benefits: 
\textbf{(1)} it can capably serve as an effective training scheme to embed semantic associations of human body parts from multiple label domains into the human representation learning process; 
\textbf{(2)} it is an extensible semantic transfer framework without predetermining the overall relations of multiple label domains, which allows continuously adding human parsing datasets to promote the training.
\textbf{(3)} the relevant modules are only used for auxiliary training and can be removed during inference, eliminating the extra reasoning cost. 
Experimental results demonstrate SST can effectively achieve promising universal human parsing performance as well as impressive improvements compared to its counterparts on three human parsing benchmarks (\textit{i.e.}, PASCAL-Person-Part, ATR, and CIHP). Code is available at \url{https://github.com/yangjie-cv/SST}.

\end{abstract}

\section{Introduction}


Human parsing, which aims to assign pixel-wise category predictions for human body parts, has played a critical role in human-oriented visual content analysis, editing, and generation, \textit{e.g.,} virtual try-on~\cite{dong2019fw}, human motion transfer~\cite{kappel2021high}, and human activity recognition~\cite{gan2016concepts}.
Prior methods have chronically been dominated by developing powerful network architectures~\cite{liang2016semantic_obj,luo2018trusted,gong2018instance,ruan2019devil,zeng2021neural}.
%
%
However, these network designs usually serve as general components for human parsing in a specific labeling system, and how to use the data from multiple label domains to further improve the accuracy of the parsing network is still an open issue.

\begin{figure*}[t]
\begin{center}
\includegraphics[width=1.0\textwidth]{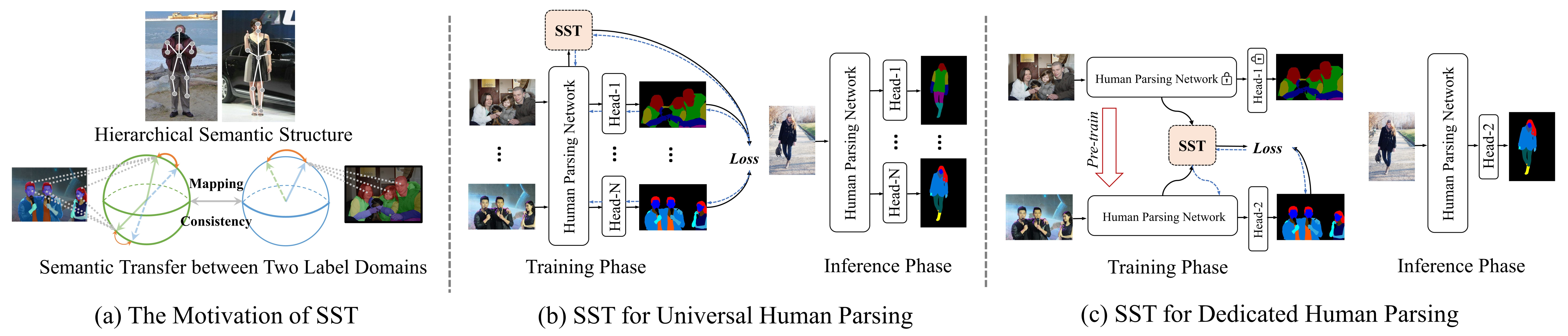}
\end{center}
\vspace{-0.5cm}
\caption{(a) The Motivation of SST: 1) Intra-Domain Hierarchical Semantic Structure (Upper): The spatial correlated two parts are connected by a white line; 2) Cross-Domain Semantic Mapping and Consistency (Lower): We regularize the semantic consistency (orange line) between original category representations (solid line) and mapped ones (dot line) for two datasets (green for dataset-1 and blue for dataset-2). (b) The Training and Inference Phase of Universal Parsing. (c) The Training and Inference Phase of Dedicated Parsing.}
\label{fig:overview}
\end{figure*}
%
Recent studies~\cite{lin2020graphonomy,he2020grapy} have proposed using multiple label domains to train a universal parsing network by constructing graph neural networks that learn semantic associations from different label domains. Graphonomy~\cite{gong2019graphonomy,lin2020graphonomy} investigates cross-domain semantic coherence through semantic aggregation and transfer on explicit graph structures. Grapy-ML~\cite{he2020grapy} extends this method by stacking three levels of human part graph structures, from coarse to fine granularity. Despite these graph-based methods consider prior human body structures, such as the spatial associations of human parts and the semantic associations of different label domains as in Fig.~\ref{fig:overview}-(a), they have limitations in their ability to dynamically add new label domains due to their dependence on pre-determined graph structures. Additionally, explicit graph reasoning incurs extra inference costs, making it impractical in some scenarios. Therefore, there is a need for a more general, scalable, and simpler manner to harness the mutual benefits of data from multiple label domains to enhance given human parsing networks.

In this work, we propose Scalable Semantic Transfer (SST), a novel training paradigm that builds plug-and-play modules to effectively embed semantic associations of human body parts from multiple label domains into the given parsing network.
The proposed SST scheme can be easily applied to two common scenarios, \textit{i.e.}, universal parsing and dedicated parsing. 
Similar to~\cite{lin2020graphonomy,he2020grapy}, \textbf{universal parsing} aims to enforce the parsing network to learn homogeneous human representation from multiple label domains with the help of the SST scheme in the \textit{training} phase. It achieves different levels of human parsing prediction for an arbitrary image in the \textit{inference phase}, as illustrated in Figure~\ref{fig:overview}-(b).
In contrast, \textbf{dedicated parsing} aims to 
optimize a parsing network for a specific label domain by employing the SST scheme to distill the semantic knowledge from other label domains in the \textit{training} phase, and finally realizes more accurate prediction for the specific label domain in the \textit{inference} phase, as shown in Fig.~\ref{fig:overview}-(c).

The proposed SST has several appealing benefits:
(1) \textit{\textbf{Generality.}} It is a general training scheme that effectively incorporates prior knowledge of human body parts into the representation learning process of a given parsing network. It enables the integration of both intra-domain spatial correlations and inter-domain semantic associations, making it suitable for training both universal and dedicated parsing networks.
%
%
%
(2) \textit{\textbf{Scalability.}} It is a scalable framework that enables the addition of new datasets for more effective training.
%
In practice, universal parsing enables the parsing network to acquire a more robust human representation by expanding the label domain pool, while dedicated parsing facilitates the transfer of rich semantic knowledge from more datasets to enhance the parsing network for a specific label domain.
%
%
(3) \textit{\textbf{Simplicity.}} All modules in SST are plug-and-play ones that are only used for auxiliary training and can be dropped out during inference, eliminating unnecessary model complexity in real-world applications.
%

The main contributions are summarized as follows:
\begin{itemize}
\item [1)] We propose a novel training paradigm called Scalable Semantic Transfer (SST) that effectively leverages the benefits of data from different label domains to train powerful universal and dedicated parsing networks.

\item [2)] We introduce three plug-and-play modules in SST that incorporate prior knowledge of human body parts into the training process of a given parsing network, which can be removed during inference.
\item [3)] Extensive experiments demonstrate the effectiveness of SST on both universal and dedicated parsing tasks, achieving impressive results on three human parsing datasets compared with the counterparts.
\end{itemize}
\begin{figure*}[t]
\begin{center}
\includegraphics[width=1\textwidth]{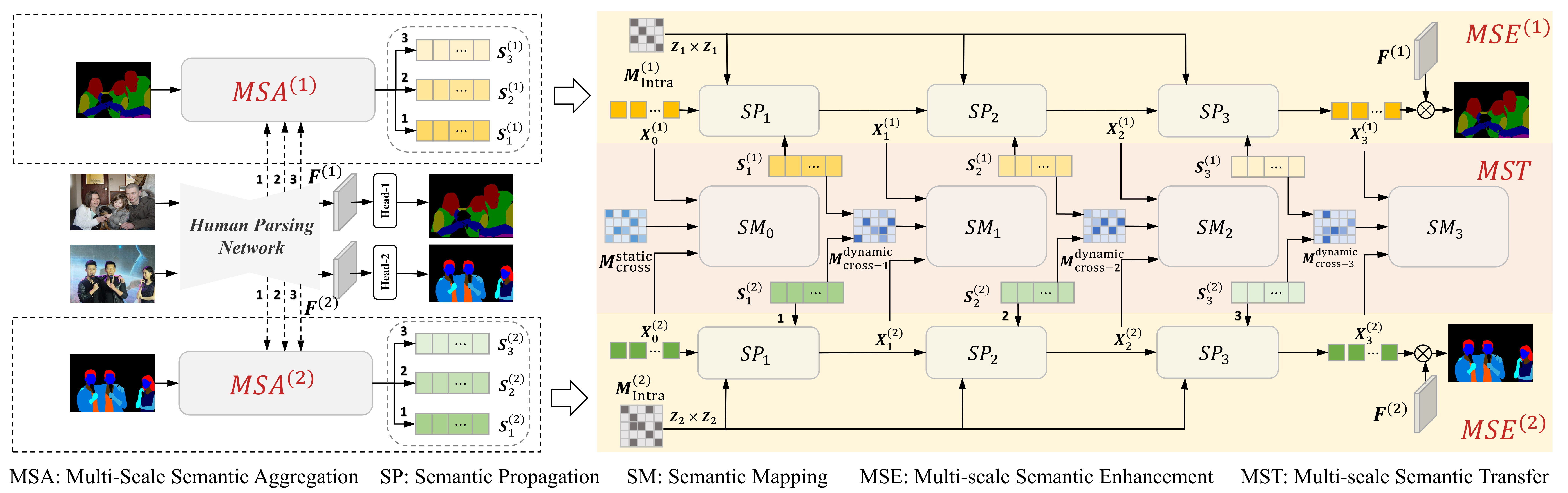}
\end{center}
\vspace{-0.5cm}
\caption{Illustration of our proposed Scalable Semantic Transfer (SST) scheme that assists to train a universal human parsing network. 
During inference, all auxiliary modules (\textit{i.e.}, MSA, MSE, and MST) are removed as shown in Fig.~\ref{fig:overview}-(b), without piling up extra computational cost. The trained parsing network can produce different levels of human parsing results given an arbitrary image as input.}
\label{fig:frmework}
\end{figure*}

\section{Related Work}
\textbf{Human Parsing.} Human parsing aims to assign pixel-wise category predictions for each human body part. Deep learning-based architectures~\cite{ruan2019devil,li2020self,zhang2020blended,zhang2020correlating,liu2022cdgnet} have achieved significant performance in a specific label domain via ingenious network designs. For example,
BGNet \cite{zhang2020blended} explores the inherent hierarchical structure of the human body,
CorrPM \cite{zhang2020correlating} discovers the spatial affinity
among feature maps from the edge, pose and parsing, and
CDGNet \cite{liu2022cdgnet} guides the network toward exploiting the intrinsic position distribution of human parts.
Since current human parsing datasets usually adopt different labeling systems, there is growing interest in exploring how to leverage the mutual benefits of data from different label domains.

Recent research efforts~\cite{gong2019graphonomy,he2020grapy} have concentrated on constructing graph structures to facilitate universal parsing across multiple label domains. Specifically, Graphonomy~\cite{gong2019graphonomy} employs intra- and inter-graphs within various datasets at different levels of granularity for multi-dataset predictions, while Graph-ML~\cite{he2020grapy} builds upon this approach by stacking multiple levels of graph structures from coarse to fine granularity for cross-domain training. 
Unlike the above existing methods that rely on graph structures for result inference, our work focuses on incorporating prior knowledge of the human body into the training process of a given parsing network, avoiding complex model reasoning.

\textbf{Attention Mechanism for Vision Modeling.} 
Recently developed attention mechanism~\cite{vaswani2017attention} for machine translation is related to our work. 
Self-attention aims to calculate the response at one position as a weighted sum of all positions in a specific context and has wide applications in vision tasks such as image classification~\cite{liu2021swin}, object detection~\cite{carion2020end}, and semantic segmentation~\cite{xie2021segformer}.
Cross-attention is another fundamental part of the attention mechanism, which can model the relations of the multiple sources tokens and is widely employed in multi-modality/multi-task/multi-domain areas. For example, 
several works~\cite{ge2022bridging, yang2022lavt} bridge the cross-modal retrieval by modeling the cross-attention within multiple modalities, while some studies~\cite{bhattacharjee2022mult, Zamir_2020_CVPR} conduct the cross-attention for multiple dense pixel prediction tasks such as segmentation and depth estimation. 

Driven by the above studies, we aim to model the semantic correlations from the intra-label domain and cross-label domain via various variants of cross-attention, which leverages prior knowledge of human body parts to guide the training process.

\section{Scalable Semantic Transfer}
%
%

Given an input image $\mathbf{I}\in  \mathbb{R}^{ H \times W \times3}$ and its label map $\mathbf{Y}$,
human parsing aims to assign each pixel with a semantic label, which is corresponding to a specific human body part.  
A popular parsing network is usually composed of a multi-stage feature extractor and a prediction head. 
The former generates a set of feature maps $ \mathbf{F} \in \mathbb{R}^{H \times W \times D}$ with the same spatial resolution as the input image, where $H$, $W$ and $D$ denote height, width, and the number of channels respectively.
The latter then converts the $ \mathbf{F}$ to categorical logit maps $\mathbf{A} \in \mathbb{R}^{H \times W \times Z}$, where $Z$ indicates the number of semantic labels for a specific dataset. 
The cross-entropy loss ${\rm{CE}(\cdot)}$ is used to train the parsing network,
\begin{equation}
\mathcal{L}_{\rm{seg}}=\frac{1}{|\Omega|}\sum_{ \mathbf{I} \in\Omega} {\rm{CE}}(\sigma(\mathbf{A}), \mathbf{Y}),
\label{eq:cross_entropy}
\end{equation}
where $\sigma(\cdot)$ is the pixel-wise softmax function and $\Omega$ indicates the training set of a specific label domain. 
In this paper, our human parsing network is a simple architecture with three components: backbone, encoder, and decoder. 
The prediction head is a simple $1\times1$ convolutional layer. 
%
%

\subsection{Universal Parsing \textbf{\textit{vs.}} Dedicated Parsing}

Given a parsing network, our proposed SST scheme aims to utilize the data from different label domains $\{\Omega_1, \Omega_2,...,\Omega_N\}$ for effective training. We apply SST to two common scenarios, \textit{i.e.}, universal parsing and dedicated parsing. The former is to enforce the given parsing network to learn homogenous human representation from different label domains and realize the multi-granularity predictions given an arbitrary image. The latter is to enhance the given parsing network for a specific label domain by distilling the semantic knowledge of other label domains.

The overview of universal parsing with the SST scheme is shown in Fig.~\ref{fig:frmework}. Here, only two datasets are used as an  illustration of SST for simplicity and more datasets can be easily extended.
We feed the images from two datasets with different labeling systems (\textit{i.e.} two label domains) into a shared parsing network and predict their pixel-level labels by using their respective prediction heads. Based on it, 
our SST scheme designs two plug-and-play modules termed \textit{Intra-Domain Multi-scale Semantic Enhancement} (MSE) and \textit{Cross-Domain Multi-scale  Semantic Transfer} (MST).
Specifically, we initialize an MSE to learn the structured label association within each label domain,
%
and initialize an MST to bridge two label domains by semantic consistency regularization.
%
%
Both MSE and MST communicate with the parsing network via the \textit{Multi-scale Semantic Aggregation} module (MSA).
After the training phase, we maintain the basic parsing network while discarding all plug-and-play modules as shown in Fig.~\ref{fig:overview}-(b), where the trained network can produce different levels of human parsing results given an arbitrary image as input.

\begin{figure}
\begin{center}
\includegraphics[width=0.46\textwidth]{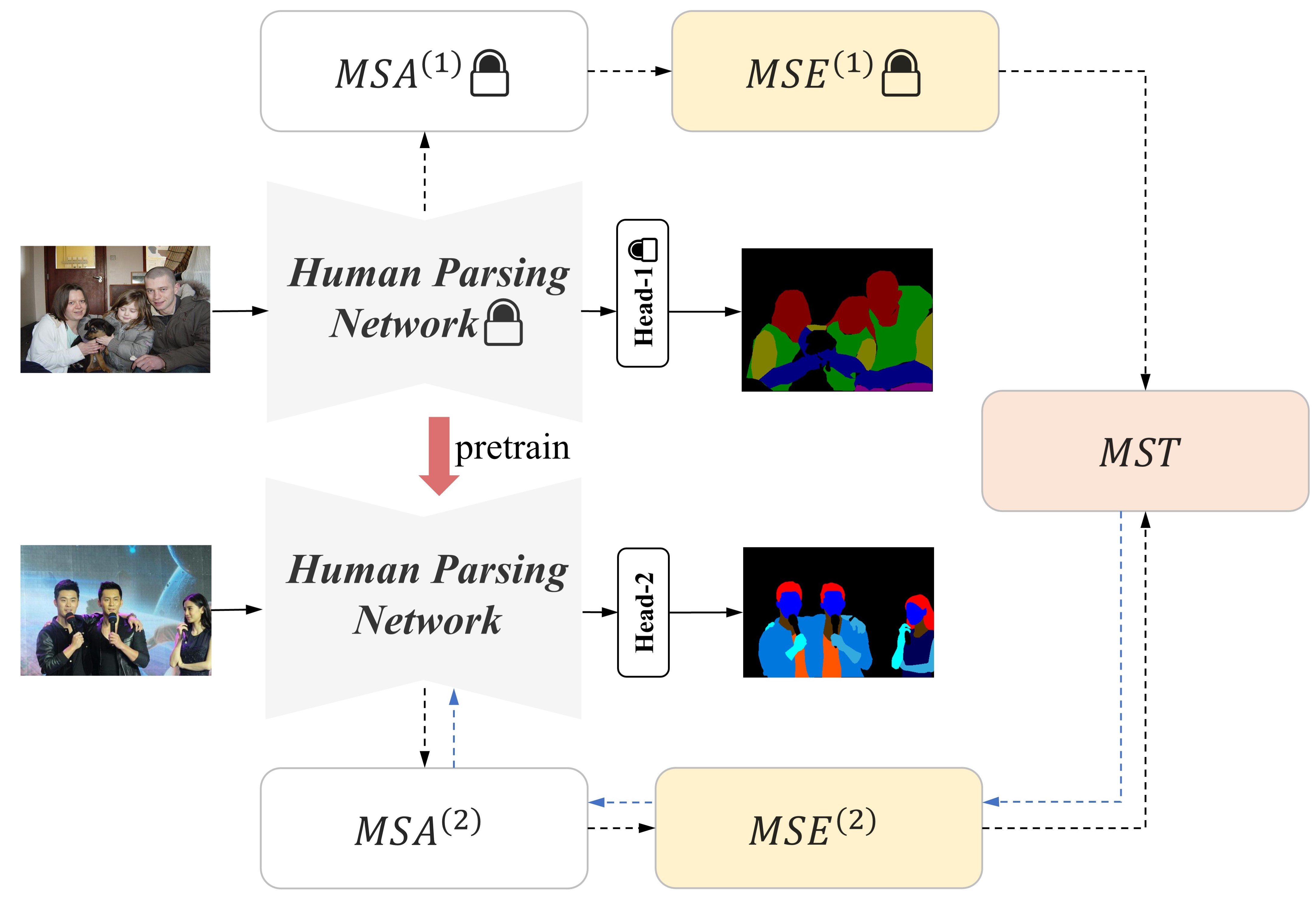}
\end{center}
\vspace{-0.5cm}
\caption{The extension of our SST scheme to dedicated parsing via a semantic knowledge distillation strategy, where the training and inference phase correspond to Fig.~\ref{fig:overview}-(c). 
}
\label{fig:scalabel}
\vspace{-0.3cm}
\end{figure}

The extension of the SST scheme for dedicated parsing is shown in Fig.~\ref{fig:scalabel}. Unlike universal parsing which applies SST to directly train the parsing network over multiple label domains, dedicated parsing adopts an ingenious training strategy via semantic knowledge distillation. Specifically, we first pre-train the given parsing network with its corresponding MSA and MSE modules using the source dataset (one or more datasets). Then, we fix the above-trained pipeline and re-train the pre-trained parsing network with random initialized MSA, MSE, and MST modules using the target dataset, where MST seamlessly bridges two label domains and
effectively distills the semantic knowledge from the source label domain to the target label domain. Finally, we remove all auxiliary modules and keep the basic network for the target label domain prediction as in Fig.~\ref{fig:overview}-(c).

In the following subsections, we will describe each module (\textit{i.e.}, MSA, MSE, and MST) in SST in detail, and give the overall objective function for the training of universal parsing and dedicated parsing respectively.



\subsection{The Components of SST}
\subsubsection{Multi-scale Semantics Aggregation (MSA)} 

For a specific label domain, we design Multi-scale Semantics Aggregation (MSA) to obtain the semantic features of each body part.
Following Grapy-ML~\cite{he2020grapy}, we adopt category-aware pooling to aggregate the features in each semantic region.
%
%
%
Specifically, we apply the multi-scale features $\mathbf{H}_l \in \mathbb{R}^{\frac{H}{4l} \times \frac{W}{4l} \times D}$,
$l\in\{1,2,3\}$ extracted from the decoder (\textit{i.e.} with resolution $1/16$, $1/8$ and $1/4$ of the input) to calculate the semantic features by average pooling, 
\begin{equation}
\mathbf{S}_{l,z}^{ave}=\frac{1}{\left| \delta_{l,z}\right|}\sum_{(i,j)\in\delta_{l,z }}\mathbf{H}_l(i,j),
\label{eq:s_ave}
\end{equation}
where $z\in \{1,2,...,Z \}$ and
$\delta_{l,z}$ indicates the semantic regions (\textit{i.e.} category mask) that are corresponding to the $z$-th category at $l$-th scale. 
$\mathbf{S}_{l,z}^{ave}$ denotes the aggregated category-wise features of $z$-th category at $l$-th scale. 
To enhance such semantic features, we further adopt max pooling instead of average pooling in Eqn. (\ref{eq:s_ave}) to calculate the category-wise features as follows,
\begin{equation}
\mathbf{S}_{l,z}^{max}=\mathop{\rm{max}}\limits_{(i,j)\in\delta_{l,z }}\mathbf{H}_l(i,j),
\label{eq:s_max}
\end{equation}
Then, we concatenate $\mathbf{S}_{l,z}^{ave}$ and $\mathbf{S}_{l,z}^{max}$ along the channel dimension and calculate the final aggregated semantic features as follows:
\begin{equation}
\mathbf{S}_{l,z}={{\rm{concat}}(\mathbf{S}_{l,z}^{ave},\mathbf{S}_{l,z}^{max})} \mathbf{W}_s,
\end{equation}
where $\mathbf{W}_s\in\mathbb{R}^{2D \times D}$ is the parameter matrix for linear projection to reduce the dimension.
%
%
The overall $l$-th scale categories' semantic features of the input image from a specific label domain can be written as the matrix form $\mathbf{S}_l=\{\mathbf{S}_{l,z}\}_{z=1}^Z$.

\subsubsection{Multi-scale Semantic Enhancement (MSE)}
Traditional semantic segmentation networks often overlook the structured nature of the human body when tackling human parsing.
Motivated by previous work~\cite{li2022deep}, we propose the Multi-scale Semantic Enhancement (MSE) module as an auxiliary training branch to learn the structured label association within a specific label domain,
where \textit{spatial association attention} is used to learn to propagate semantic information between two adjacent human body regions.



%
As illustrated in the right of Fig.~\ref{fig:frmework}, we randomly initialize the learnable semantic embedding $\mathbf{X}_{0}$ for a specific label domain, which can be considered as the learnable category representations of the corresponding dataset.
In practice, we employ the $\mathbf{X}_{0}$ as the input of the first semantic propagation layer (SP) and perform the cross-attention operation with category features at the first scale ($\textbf{S}_1$).
The output of the SP is the updated category representations and we feed it into the next SP to do the interaction with the category features at the next scale.
In general, we can formulate the processing in $l$-th SP layer as follows, 
\begin{equation}
\mathbf{X}_{l}={\rm{softmax}}(\frac{\mathbf{Q}_l\mathbf{K}_l^\mathsf{T}}{\sqrt{D}})\mathbf{V}_l+\mathbf{X}_{l-1},
\label{eq:intra_cross}
\end{equation}
where $l\in\{1,2,3\}$ is the scale/layer index, 
$\mathbf{X}_{l} \in \mathbb{R}^{Z \times D}$ refers to the category representations. 
%
%
$Z$ indicates the total number of semantic categories of the dataset, and $D$ is the feature dimension.
Here we have $\mathbf{Q}_{l}=f_Q(\mathbf{X}_{l-1}) $, 
$\mathbf{K}_l = f_K(\mathbf{S}_{l}) $, 
and $\mathbf{V}_l = f_V(\mathbf{S}_{l})$. 
All of $f_Q(\cdot)$, $f_K(\cdot)$ and $f_V(\cdot)$ are linear transformations with the output dimension ${Z \times D}$.
To embed the prior knowledge of human body part connections, we further propose spatial association attention and reformulate Eqn.~(\ref{eq:intra_cross}) as,
\begin{equation}
\mathbf{X}_{l}={\rm{softmax}}(\frac{\mathbf{Q}_l\mathbf{K}_l^\mathsf{T}}{\sqrt{D}})\odot\mathbf{M}_{\rm{intra}} \mathbf{V}_l+\mathbf{X}_{l-1},
\label{eq:intra_cross_mask}
\end{equation}
where $\mathbf{M}_{\rm{intra}}$ is a binary matrix to present the prior knowledge of relationships between different categories within the label domain~\cite{gong2019graphonomy}, 
where the element with the value $1$ in the matrix indicates that the corresponding two semantic categories have spatial connectivity on the human body, and $0$ otherwise.


After passing through all of the SP layers, we conduct the dot-product operation between image category representation $\mathbf{X}_L$ at the last scale and feature map $\mathbf{F}$ to calculate the auxiliary mask as follows,
\begin{equation}
\mathbf{A}_{\rm{aux}}=\mathbf{F} \otimes \mathbf{X}_L^{\mathsf{T}},
\label{eq:aux}
\end{equation}
where we set $L=3$ in this paper and $\mathbf{A}_{\rm{aux}}\in \mathbb{R}^{H \times W \times Z}$ is also supervised by using cross-entropy loss with the same form as Eqn.~(\ref{eq:cross_entropy}), denoted as $\mathcal{L}_{\rm{aux}}$.

\subsubsection{Multi-scale Semantic Transfer (MST)}
\label{sec:MST}
%
%

Even though different label domains may have distinct part labels, there are often explicitly structured correlations among them that can be leveraged. For instance, the label \texttt{head} in the Pascal-Person-Part dataset~\cite{chen2014detect} actually encompasses {\ttfamily hat}, {\ttfamily hair}, {\ttfamily face} in CIHP dataset~\cite{gong2018instance}, as depicted in Fig.~\ref{fig:overview}-(a).
Motivated by this, we propose Cross-Domain Multi-scale Semantic Transfer (MST) module
for
cross-label domain semantic mapping and consistency regularization at different scales.

%

\textbf{Multi-scale Semantic Mapping.} 
%
%
We introduce the semantic mapping layer (SM) and propose two variants of cross-attention termed \textit{static semantic association attention} and
\textit{dynamic semantic association attention}, 
to map the category representations from one semantic space to another.

As illustrated in the right of Fig.~\ref{fig:frmework}, we employ the superscript to indicate the datasets belonging to different label domains. 
Then the global learnable category representations of two datasets can be presented as  $\mathbf{X}_0^{(1)} \in \mathbb{R}^{Z_1 \times D}$ and $\mathbf{X}_0^{(2)}\in \mathbb{R}^{Z_2 \times D}$.
The proposed static semantic association attention operation in the \textit{zeroth} SM layer can be formulated as follows,
\begin{equation}
\mathbf{X}_{0}^{(1) \rightarrow (2)}={\rm{softmax}}(\frac{\hat{\mathbf{Q}}_{0}^{(2)} \hat{\mathbf{K}}_{0}^{{(1)}\mathsf{T}}}{\sqrt{D}})\odot\mathbf{M}_{\rm{cross}}^{\rm{static}} \hat{\mathbf{V}}_{0}^{(1)},
\label{eq:global_cross}
\end{equation}
%
%
where $\mathbf{M}_{\rm{cross}}^{\rm{static}}\in \mathbb{R}^{Z_2 \times Z_1}$ is a static matrix whose elements denote the similarities of labels calculated by the cosine distances of their word embeddings~\cite{gong2019graphonomy}. 
$\hat{\mathbf{Q}}_{0}^{(2)}$ is calculated as the linear projection of $\mathbf{X}_0^{(2)}$.
Similarly, both $\hat{\mathbf{K}}_{0}^{(1)}$ and $\hat{\mathbf{V}}_{0}^{(1)}$ are obtained by $\mathbf{X}_0^{(1)}$.
Here $\mathbf{X}_{0}^{(1) \rightarrow (2)} \in \mathbb{R}^{Z_2 \times D}$  indicates the global category representation mapping results from the first label domain to the second one.
Symmetrically, we can obtain $\mathbf{X}_{0}^{(2) \rightarrow (1)} \in \mathbb{R}^{Z_1 \times D}$ in the same way.

Since $\mathbf{M}_{\rm{cross}}^{\rm{static}}$ is essentially semantic prior, where its values are calculated based on the word embedding of each label, 
%
it is insufficient to reflect the semantic association between the specific image pairs from different domains.
To address this limitation, we conduct the dynamic semantic association attention operation in the subsequent SM layers.
Given the category representations  $\mathbf{X}_l^{(1)} \in \mathbb{R}^{Z_1 \times D}$ and $\mathbf{X}_l^{(2)}\in \mathbb{R}^{Z_2 \times D}$ of two images extracted from $l$-th SP layer, 
the dynamic semantic association attention in the $l$-th SM layer can be formulated as follows,
\begin{equation}
\begin{aligned}
\mathbf{X}_{l}^{(1) \rightarrow (2)}&={\rm{softmax}}(\frac{\hat{\mathbf{Q}}_{l}^{(2)}\hat{\mathbf{K}}_{l}^{{(1)}\mathsf{T}}}{\sqrt{D}})\odot {\mathbf{M}_{ {\rm{cross}}~l}^{\rm{dynamic}} } \hat{\mathbf{V}}_{l}^{(1)}, \\
\mathbf{M}_{{\rm{cross}}~l}^{\rm{dynamic}} &= \mathbf{S}_{l}^{(2)}\mathbf{W}_{l}^{(2)}(\mathbf{S}_{l}^{(1)}\mathbf{W}_{l}^{(1)})^\mathsf{T},
\end{aligned}
\label{eq:local_cross}
\end{equation}
where $\mathbf{M}_{{\rm{cross}}~l}^{\rm{dynamic}}\in \mathbb{R}^{Z_2 \times Z_1} $ is the dynamic adjacency matrix whose elements measure the category features' similarities between $\mathbf{S}_{l}^{(2)}$ and $\mathbf{S}_{l}^{(1)}$, which are extracted from $l$-th scale of MSA module. 
Both $\mathbf{W}_{l}^{(1)}$ and $\mathbf{W}_{1}^{(2)}$ are the parameter matrices for linear projection.
Same as Eqn.~(\ref{eq:global_cross}), $\hat{\mathbf{Q}}_{l}^{(2)}$, $\hat{\mathbf{K}}_{l}^{(1)}$ and $\hat{\mathbf{V}}_{l}^{(1)}$ are calculated by using $\mathbf{X}_l^{(2)}$ and $\mathbf{X}_l^{(1)}$ respectively.
%
%
Similarly, we can also calculate the semantic mapping from the second label domain to the first one, 
and obtain  $\mathbf{X}_{l}^{(2) \rightarrow (1)} \in \mathbb{R}^{Z_1 \times D}$.
%

%

\textbf{Multi-scale Semantic Consistency Regularization.}
Motivated by the assumption that mapping representations and original representations in the same semantic space should be consistent,
%
we propose Semantic Consistency Regularization (SCR) at both the dataset-level and image-level to bridge two label domains for semantic knowledge sharing.
Specifically, we symmetrically align the global category representations $\mathbf{X}_{0}^{(1)}, \mathbf{X}_{0}^{(2)}$ and the corresponding transformed ones $\mathbf{X}_{0}^{(2) \rightarrow (1)}, \mathbf{X}_{0}^{(1) \rightarrow (2)}$ at dataset-level as follows,
%
%
%
%
\begin{equation}
\begin{aligned}
        \mathcal{L}_{\rm{SCR}}^{\rm{dataset}} 
        &= \frac{1}{Z_1}\sum_{z_1=1}^{Z_1} (1-\frac{\mathbf{X}_{0,z_1}^{(1)~\mathsf{T}} \mathbf{X}_{0,z_1}^{(2) \rightarrow (1)}}{\Vert \mathbf{X}_{0,z_1}^{(1)} \Vert \cdot \Vert \mathbf{X}_{0,z_1}^{(2) \rightarrow (1)} \Vert})\\
        &+ \frac{1}{Z_2}\sum_{z_2=1}^{Z_2} (1-\frac{\mathbf{X}_{0,z_2}^{(2)~\mathsf{T}} \mathbf{X}_{0,z_2}^{(1) \rightarrow (2)}}{\Vert \mathbf{X}_{0,z_2}^{(2)} \Vert \cdot \Vert \mathbf{X}_{0,z_2}^{(1) \rightarrow (2)} \Vert}),
\end{aligned}
\label{eq:mcr_dataset}
\end{equation}
Also, the image category representations $\mathbf{X}_{l}^{(1)}, \mathbf{X}_{l}^{(2)}$ and the corresponding transformed ones $\mathbf{X}_{l}^{(2) \rightarrow (1)}, \mathbf{X}_{l}^{(1) \rightarrow (2)}$ can be aligned at image-level as follows,
\begin{equation}
\begin{aligned}
        \mathcal{L}_{\rm{SCR}}^{\rm{img}}&= \frac{1}{Z_1}\sum_{l=1}^L\sum_{z_1=1}^{Z_1} (1-\frac{\mathbf{X}_{l,z_1}^{(1)~\mathsf{T}} \mathbf{X}_{l,z_1}^{(2) \rightarrow (1)}}{\Vert \mathbf{X}_{l,z_1}^{(1)} \Vert \cdot \Vert \mathbf{X}_{l,z_1}^{(2) \rightarrow (1)} \Vert})\\
        &+\frac{1}{Z_2}\sum_{l=1}^L\sum_{z_2=1}^{Z_2}
        (1-\frac{\mathbf{X}_{l,z_2}^{(2)~\mathsf{T}} \mathbf{X}_{l,z_2}^{(1) \rightarrow (2)}}{\Vert \mathbf{X}_{l,z_2}^{(2)} \Vert \cdot \Vert \mathbf{X}_{l,z_2}^{(1) \rightarrow (2)} \Vert}),
\end{aligned}
\label{eq:mcr_img}
\end{equation}
Intuitively, the benefits of image-level transformation and consistency learning in Eqn.~(\ref{eq:local_cross}) and Eqn.~(\ref{eq:mcr_img}) are from two aspects.
%
First, the dynamic adjacency matrix can stabilize the training process when semantic regions of certain categories may not appear in the image.
Second, the dynamic adjacency matrix is generated by using any two images from different label domains, allowing a large number of sample pairs for consistency learning.

\subsection{The Overall Objective Function}
\textbf{Universal Parsing.}
The training loss of one-step universal parsing is as follows,
\begin{equation}
\begin{aligned}
\mathcal{L}_{u}&=\alpha (\mathcal{L}_{\rm{seg}}^{(1)}+\mathcal{L}_{\rm{seg}}^{(2)})+\beta(\mathcal{L}_{\rm{aux}}^{(1)}+\mathcal{L}_{\rm{aux}}^{(2)}) \\
& +\lambda (\mathcal{L}_{\rm{SCR}}^{\rm{dataset}}+ \mathcal{L}_{\rm{SCR}}^{\rm{img}}),
\end{aligned}
\end{equation}

\textbf{Dedicated Parsing.}
The training loss of two-step dedicated parsing can be written as follows,
\begin{equation}
\begin{aligned}
\mathcal{L}_{d}^{\rm{first}}&=\alpha \mathcal{L}_{\rm{seg}}^{(1)}+\beta\mathcal{L}_{\rm{aux}}^{(1)},\\
\mathcal{L}_{d}^{\rm{second}}&=\alpha \mathcal{L}_{\rm{seg}}^{(2)}+\beta\mathcal{L}_{\rm{aux}}^{(2)} +\lambda( \mathcal{L}_{\rm{SCR}}^{\rm{dataset}}+ \mathcal{L}_{\rm{SCR}}^{\rm{img}}),
\end{aligned}
\end{equation}
where $\mathcal{L}_{\rm{seg}}^{(1)}, \mathcal{L}_{\rm{seg}}^{(2)}$ indicate the main segmentation loss for two datasets,
and $\mathcal{L}_{\rm{aux}}^{(1)}, \mathcal{L}_{\rm{aux}}^{(2)}$ are calculated by using Eqn.~(\ref{eq:aux}) over all of the samples of two datasets.
%




\section{Experiments}

\subsection{Experiment Settings}

\textbf{Datasets and Evaluation Metric.}
We evaluate our proposed method on three human parsing benchmarks with different labels granularities, including
PASCAL-Person-Part dataset~\cite{chen2014detect} for $7$ labels, ATR dataset~\cite{liang2015human} for $18$ labels,
Crowd Instance-Level Human Parsing (CIHP) dataset~\cite{gong2018instance} for 20 labels.
We employ the same evaluation metrics as Grapy-ML\cite{he2020grapy} for a fair comparison, \textit{i.e.} mean accuracy and mean intersection over union (mIoU).

%

%
%



\textbf{Human Parsing Network.}
We adopt ResNet-50 and ResNet-101 pre-trained on ImageNet as our backbone. Following \cite{cheng2022masked}, we use multi-scale deformable attention Transformer \cite{zhu2020deformable} as our encoder and three successive transformers~\cite{liu2021swin} with up-sampling operation as our decoder. The prediction head is a simple $1\times1$ convolutional layer, which varies according to the different datasets.
%

\textbf{Implementation Details.} 
Following Grapy-ML\cite{he2020grapy} and Graphonomy\cite{gong2019graphonomy}, the training images were augmented by a random resize from $0.5$ to $2$, $512\times 512$ cropping and horizontal flipping.
We trained all the models for $150$ epochs with an initial learning rate of $1e-4$. We adopt a simple learning policy that drops the learning rate by multiplying $0.1$ at the $125$-th epoch. The loss coefficients $\alpha$, $\beta$ and $\lambda$ are set as $10$, $1$ and $5$.
All experiments are implemented on NVIDIA Tesla A100 GPUs and the batch size is set as $4$. For multi-dataset training, images from multiple datasets can be processed in a batch, which will be split only when passed through dataset-specific layers. During the inference,
we removed all of the auxiliary branches and maintained the original parsing network without adding extra parameters. Following Grapy-ML\cite{he2020grapy} and Graphonomy\cite{gong2019graphonomy}, we averaged all predictions from horizontal flipped and multi-scale inputs as the final prediction. 


\begin{figure*}[h]	
\centering
 	{
 		\begin{minipage}[t]{1\linewidth}
 			\centering         
 			\includegraphics[width=1\linewidth]{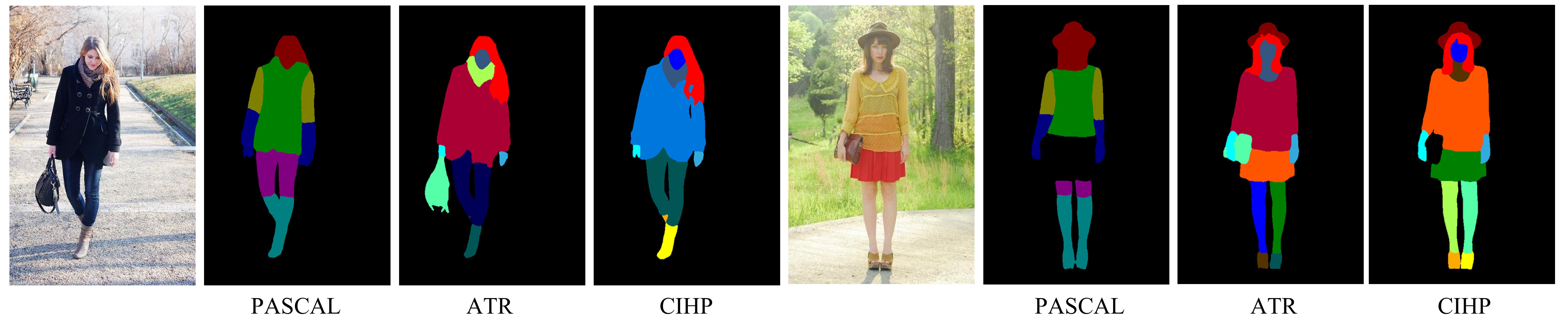}  
 		\end{minipage}
 	} 
\vspace{-0.6cm}
\caption{Illustrative results of our universal parsing network with SST scheme. Given an image, our model can produce precise results for different levels of human parsing tasks by switching the corresponding prediction head (\textit{i.e.}, PASCAL-Person-Part, ATR, and CIHP).}
\label{fig:universal} 
\vspace{-0.4cm}
\end{figure*}

\begin{table}
\centering
\resizebox{0.8\linewidth}{!}{
\begin{tabular}{l|c|cc} 
\toprule [1.5 pt]
Method & Backbone &  mIoU ($\%$)\\
\hline
PGN \cite{gong2018instance} &Xception    &  68.40 \\
Deeplab V3+ \cite{chen2018encoder} & Xception    & 67.60 \\
Deeplab V3+ \cite{chen2018encoder} & ResNet-101    & 67.84 \\
Learning \cite{wang2019learning} & ResNet-101  & 70.76\\
DPC \cite{chen2018searching} & Xception  & 71.34\\
SNT \cite{ji2020learning} & ResNet-101   & 71.59\\
GPM \cite{he2020grapy} &Xception & 69.50 \\
NPPNet  \cite{zeng2021neural} & - & 71.73\\
SCHP \cite{li2020self} &ResNet-101 & 71.56\\ \rowcolor{gray!5}
Single-dataset &ResNet-50  &  67.15\\\rowcolor{gray!5}
Single-dataset &ResNet-101  &  68.32\\
\hline
Grapy-ML \cite{he2020grapy} &Xception &  71.65 \\ 
Graphonomy \cite{gong2019graphonomy} &Xception & 71.14\\  \rowcolor{gray!10}
Multi-datasets& ResNet-50 &  70.40\color{Red}$\uparrow_{3.25}$ \\ \rowcolor{gray!10}
SST (Universal) & ResNet-50 & 72.25\color{Red}$\uparrow_{5.10}$ \\\rowcolor{gray!10} 
SST (Transfer-CIHP) & ResNet-50 & 73.49\color{Red}$\uparrow_{6.34}$ \\\rowcolor{gray!10} 
SST (Transfer-ATR\&CIHP) & ResNet-50 & 73.87\color{Red}$\uparrow_{6.72}$ \\\rowcolor{gray!20} 
Multi-datasets & ResNet-101 &  71.41\color{Red}$\uparrow_{3.09}$\\ \rowcolor{gray!20}
SST (Universal) & ResNet-101 & 73.13\color{Red}$\uparrow_{4.81}$ \\\rowcolor{gray!20}
SST (Transfer-CIHP) & ResNet-101 & 74.52\color{Red}$\uparrow_{6.20}$  \\ \rowcolor{gray!20}
SST (Transfer-ATR\&CIHP) & ResNet-101 &  \textbf{74.96}\color{Red}$\uparrow_{6.64}$ \\

  \bottomrule[1.5 pt]
  \end{tabular}}
    \vspace{-0.2cm}
\caption{Comparison on PASCAL-Person-Part Dataset.}
  \label{table:Pascal}
  \vspace{-0.3cm}
\end{table}

\subsection{Universal Human Parsing}
Fig.~\ref{fig:universal} illustrates that SST can effectively leverage data from various label domains to train a robust universal parsing network, referred to as \textit{SST (Universal)}. To highlight the advantages conferred by SST, we additionally train a parsing network using three distinct datasets without employing SST, denoted as \textit{Multi-datasets}.
For a fair comparison, we evaluate the performance of our \textit{Single-dataset} model against previous methods trained on individual datasets. Furthermore, we compare our multi-datasets model with Grapy-ML \cite{he2020grapy} and Graphonomy \cite{gong2019graphonomy}.

\textbf{PASCAL-Person-Part.}
\label{PASCAL-Person-Part}
Our \textit{Single-dataset} model attains $67.15\%$ mIoU and $68.32\%$ mIoU with ResNet-50 and ResNet-101 backbones as in Tab.~\ref{table:Pascal}, respectively. However, it underperforms compared to other methods trained on a single dataset. This is primarily due to the limited training samples in the PASCAL-Person-Part dataset, which comprises only $1,717$ samples, thereby constraining the optimization of our parsing network.
Moreover, the \textit{Multi-datasets} model achieves competitive results with $70.40\%$ mIoU and $71.41\%$ mIoU when the training dataset is directly expanded. Notably, our SST approach improves mIoU by $1.85\%$ and $1.72\%$ based on the \textit{Multi-datasets} model under the two backbones, respectively, surpassing the performance of Grapy-ML \cite{he2020grapy} and Graphonomy \cite{lin2020graphonomy}.

\begin{table}
\centering
\resizebox{\linewidth}{!}{
\begin{tabular}{l|c|cc} 
\toprule [1.5 pt]
Method & Backbone &  Mean Acc. ($\%$) & mIoU ($\%$)\\
\hline
JPPNet \cite{liang2018look} &ResNet-101 & - & 54.45  \\
Deeplab V3+ \cite{chen2018encoder}  &Xception & 84.07&76.52\\
GPM  \cite{he2020grapy} & Xception& 84.44& 76.97\\\rowcolor{gray!5}
Single-dataset & ResNet-50 &  87.10 & 79.15\\\rowcolor{gray!5}
Single-dataset & ResNet-101 &  87.56 & 79.52\\
\hline
Grapy-ML \cite{he2020grapy}& Xception &85.22 &77.88  \\
Graphonomy* \cite{lin2020graphonomy}& Xception&83.98& 76.35 \\\rowcolor{gray!10}
Multi-datasets & ResNet-50& 87.41\color{Red}$\uparrow_{0.31}$ &79.39\color{Red}$\uparrow_{0.24}$  \\\rowcolor{gray!10}
SST (Universal) &ResNet-50 &  87.95\color{Red}$\uparrow_{0.85}$  & 79.97\color{Red}$\uparrow_{0.82}$\\\rowcolor{gray!10}
SST (Transfer-CIHP) &ResNet-50 &  88.54\color{Red}$\uparrow_{1.44}$ & 81.05\color{Red}$\uparrow_{1.90}$\\\rowcolor{gray!10}
SST (Transfer-PASCAL\&CIHP) &ResNet-50 & 88.78\color{Red}$\uparrow_{1.68}$   &  81.36\color{Red}$\uparrow_{2.21}$ \\\rowcolor{gray!20}
Multi-datasets & ResNet-101& 87.82\color{Red}$\uparrow_{0.26}$ &79.82\color{Red}$\uparrow_{0.30}$ \\\rowcolor{gray!20}
SST (Universal)  &ResNet-101 &  88.12\color{Red}$\uparrow_{0.56}$  & 80.50\color{Red}$\uparrow_{0.98}$ \\  \rowcolor{gray!20}
SST (Transfer-CIHP)  &ResNet-101 &  88.98\color{Red}$\uparrow_{1.42}$  & 81.77\color{Red}$\uparrow_{2.25}$ \\ \rowcolor{gray!20}
SST (Transfer-PASCAL\&CIHP)  &ResNet-101 &  \textbf{89.26}\color{Red}$\uparrow_{1.70}$   & \textbf{82.15}\color{Red}$\uparrow_{2.63}$  \\

\bottomrule[1.5 pt]
\end{tabular}}
  \vspace{-0.2cm}
\caption{Comparison on ATR Dataset. The result of Graphonomy* is reproduced by Grapy-ML~\cite{he2020grapy}.}
  \label{tbl:ATR}
  \vspace{-0.1cm}
\end{table}
\textbf{ATR.}
As shown in Tab.~\ref{tbl:ATR}, our \textit{Single-dataset} model achieves comparable performance with JPNet \cite{liang2018look}, Deeplab V3+ \cite{chen2018encoder}, and GPM \cite{he2020grapy}, as the size of the training set increases from $1,717$ in PASCAL-Person-Part to $16,000$ in ATR. Moreover, our universal model trained with SST exhibits a consistent improvement in performance over the \textit{Multi-dataset} model for different backbones.

\textbf{CIHP.}
Tab.~\ref{tbl:CIHP} demonstrates that our \textit{Single-dataset} model outperforms state-of-the-art methods that are also trained with a single dataset. This is due to the benefits of using the large-scale training set provided by CIHP, which enhances the optimization of our parsing network. Furthermore, our SST scheme improves the performance of our \textit{Multi-dataset} model, achieving a $63.35\%$ mIoU.

\textbf{Semantic Transfer Strategies over Multiple Datasets.} Sec.~\ref{sec:MST} outlines the process of semantic transfer between two datasets. In our experiments, we utilize three datasets to verify the extensibility of SST, where we build multiple pairwise relationships (\textit{i.e.}, $3$) in a full-connection manner for semantic transfer between any two datasets, denoted as \textbf{Full} in Tab.~\ref{tbl:strategies}. Considering reducing the training cost for more datasets, we design a progressive connection from coarse to fine, denoted as \textbf{Progressive}. It follows the chain rule to connect the coarsest-grained dataset to the most refined dataset step by step (\textit{e.g.}, $2$ pairs for $3$ datasets), which can reduce the large training cost when using a considerable number of datasets. The results show that removing some intermediate pairwise relationships may result in a slight performance drop, which is acceptable when considering the reduction in training costs.

\begin{table}
\centering
\resizebox{\linewidth}{!}{
\begin{tabular}{l|c|cc} 
\toprule [1.5 pt]
Method & Backbone & Mean Acc. ($\%$) & mIoU ($\%$)\\
\hline
PGN \cite{gong2018instance} & Xception  & 64.22 & 55.80  \\
Deeplab V3+\cite{chen2018encoder} & Xception &65.06&57.13\\
M-CE2P \cite{ruan2019devil} & ResNet-101  &- & 59.50\\
Parsing R-CNN \cite{yang2019parsing} & -  &- & 59.80\\
BraidNet \cite{liu2019braidnet} & - &- & 60.62\\
PCNet \cite{zhang2020part} &ResNet-101 &67.05 & 61.05\\
GPM \cite{he2020grapy} &Xception &68.95 & 60.36\\ \rowcolor{gray!5}
Single-dataset&  ResNet-50& 73.18 & 62.51\\\rowcolor{gray!5}
Single-dataset&  ResNet-101& 73.56 & 62.97\\
\hline
Grapy-ML \cite{he2020grapy} &Xception & 68.97 &60.60 \\
Graphonomy \cite{lin2020graphonomy} &Xception & 66.65 & 58.58 \\\rowcolor{gray!10}
Multi-datasets &ResNet-50 &  73.36\color{Red}$\uparrow_{0.18}$ & 62.74\color{Red}$\uparrow_{0.23}$ \\\rowcolor{gray!10}
SST (Universal)& ResNet-50 & 73.97\color{Red}$\uparrow_{0.79}$  &  63.35\color{Red}$\uparrow_{0.84}$ \\\rowcolor{gray!10}
SST (Transfer-PASCAL)& ResNet-50 & 74.16\color{Red}$\uparrow_{0.98}$  &  63.57\color{Red}$\uparrow_{1.06}$ \\ \rowcolor{gray!10}
SST (Transfer-PASCAL\&ATR)& ResNet-50 & 74.29\color{Red}$\uparrow_{1.11}$ & 63.86\color{Red}$\uparrow_{1.35}$ \\ \rowcolor{gray!20}
Multi-datasets &ResNet-101 & 73.83\color{Red}$\uparrow_{0.27}$ & 63.18\color{Red}$\uparrow_{0.21}$ \\\rowcolor{gray!20}
SST (Universal) & ResNet-101 & 74.26\color{Red}$\uparrow_{0.70}$ &  63.65\color{Red}$\uparrow_{0.68}$ \\ \rowcolor{gray!20}
SST (Transfer-PASCAL) & ResNet-101 & 74.42\color{Red}$\uparrow_{0.86}$ &  63.89\color{Red}$\uparrow_{0.92}$ \\ \rowcolor{gray!20}
SST (Transfer-PASCAL\&ATR) & ResNet-101 & \textbf{74.56}\color{Red}$\uparrow_{1.00}$ & \textbf{64.32}\color{Red}$\uparrow_{1.34}$  \\
  \bottomrule[1.5 pt]  \end{tabular}}
    \vspace{-0.2cm}
\caption{Comparison on CIHP Dataset.}
  \label{tbl:CIHP}
\vspace{-0.1 cm}
\end{table}

\begin{table}
\centering
\resizebox{0.8\linewidth}{!}{
\begin{tabular}{c|c|c|c} 
\toprule [1 pt]
Strategies & Backbone & Dataset &mIoU ($\%$) \\
\hline
Progressive &ResNet-50 & PASCAL & 72.03\\
Full &ResNet-50 & PASCAL  &  \textbf{72.25}\\ \hline
Progressive &ResNet-50 & ATR &  79.64\\
Full &ResNet-50 & ATR  & \textbf{79.97} \\ \hline
Progressive &ResNet-50 & CIHP & 63.19 \\
Full &ResNet-50 & CIHP  &  \textbf{63.35} \\
\bottomrule[1pt]
\end{tabular}}
  \vspace{-0.2cm}
\caption{Impact on different strategies of semantic transfer over multiple datasets in our SST scheme.}
  \label{tbl:strategies}
  \vspace{-0.4cm}
\end{table}

\subsection{Dedicated Human Parsing} SST can enhance the parsing network for a specific label domain by transferring the distilled semantic knowledge from other domains.
We conduct three kinds of experiments (\textit{i.e.}, \textit{Fine-to-Coarse Transfer}, \textit{Coarse-to-Fine Transfer}, and \textit{Universal-to-Specific Transfer}) to extensively verify the effectiveness of the transfer between different granular label domains. Furthermore, we exploit an appealing merit of such transfer, which can alleviate the need for heavy annotated re-training data.

\begin{table}
\centering
\resizebox{0.8\linewidth}{!}{
\begin{tabular}{c|c|c|c} 
\toprule [1 pt]
Re-training data & Backbone &  Finetune  & SST \\
\hline
50$\%$ &ResNet-50 &  69.23  & 71.81 \\
80$\%$ &ResNet-50 &  70.85 &  72.96 \\
100$\%$  &ResNet-50 & 71.52 & \textbf{73.49} \\ 
\bottomrule[1pt]
\end{tabular}}
  \vspace{-0.2cm}
\caption{The performance of transferring the model pre-trained on the CIHP dataset to PASCAL-Person-Part dataset without (\textit{i.e.}, finetune) or with SST, in terms of mIoU ($\%$).}
  \label{tbl:pascal tranfer}
  \vspace{-0.3cm}
\end{table}

\textbf{Fine-to-Coarse Transfer.} We pre-train the model on the fine-grained CIHP dataset and then adapt it to the PASCAL-Person-Part dataset or ATR dataset using our SST scheme. Tab.~\ref{table:Pascal} and Tab.~\ref{tbl:ATR} report the fine-to-coarse results for the PASCAL-Person-Part dataset and ATR dataset, respectively. Our results show that
the re-trained model using SST for the target dataset outperforms the universal model with SST. This improvement is because training a universal model needs to handle the discrepancy of label granularity across different datasets. Instead, re-training with a new dataset only requires updating semantic knowledge based on the model to be adapted.


\textbf{Coarse-to-Fine Transfer.} We pre-trained a model on the high-level annotated PASCAL-Person-Part dataset, and transferred it to the fine-grained CIHP dataset using coarse-to-fine transfer learning. Our results, presented in Tab.~\ref{tbl:CIHP}, demonstrate that leveraging high-level semantic representations from PASCAL-Person-Part enhances the semantic learning of CIHP, achieving a promising $63.57\%$ mIoU.

\textbf{Universal-to-Specific Transfer.} Given three datasets, we utilize any two datasets to pre-train a universal parsing network using SST. Then we adapt the trained universal parsing network to the remaining dataset. This strategy achieves the best results on three datasets in Tab.~\ref{table:Pascal}, Tab.~\ref{tbl:ATR}, and Tab.~\ref{tbl:CIHP}. It shows that the learned homogeneous human representations via SST can be effectively leveraged to improve parsing performance in other specific label domains. 

\textbf{The Number of Re-training Data.} 
We experiment with different amounts of re-training data to transfer a model pre-trained on the CIHP dataset to the PASCAL-Person-Part dataset. Specifically, we randomly sample varying amounts of annotated data from the training set of PASCAL-Person-Part to re-train the model, and evaluate its performance on the entire test set. Surprisingly, as shown in Tab.~\ref{tbl:pascal tranfer}, SST significantly reduces the required re-training data to $50\%$ while still achieving better results than direct fine-tuning with all data.
Furthermore, our SST scheme with $100\%$ re-training data outperforms the fine-tuning model by $1.97\%$ mIoU, demonstrating that it can bridge different label domains by learning their semantic associations for effective semantic knowledge transfer.

\subsection{Ablation Studies for SST Scheme}
We conduct ablation studies to validate the effectiveness of key components in SST on the PASCAL-Person-Part dataset under the ResNet-50 backbone in Tab.~\ref{table:ablation}.

\textbf{Intra-Domain Multi-scale Semantic Enhancement (MSE).} MSE aims to enforce the network to learn the structured label association within a specific label domain by an auxiliary loss $\mathcal{L}_{\rm{aux}}$. In practice, we introduce a binary matrix $\mathbf{M}_{\rm{intra}}$ to encode the prior human body knowledge into such process, as in Eqn.~(\ref{eq:intra_cross_mask}). From Tab.~\ref{table:ablation}, our MSE improves mIoU by $0.62\%$ compared with the \textit{Multi-datasets} model, $1^\dag$ \textit{vs.} $3^\dag$. Further, we compare the results of the MSE with and without $\mathbf{M}_{\rm{intra}}$, $2^\dag$ \textit{vs.} $3^\dag$, which shows that the prior human body knowledge can further enhance the semantic propagation via the MSE module during training.

\textbf{Cross-Domain Multi-scale Semantic Transfer (MST).}
MST bridges the different label domains by $\mathcal{L}_{\rm{SCR}}^{\rm{dataset}}$ in Eqn.~(\ref{eq:mcr_dataset}) and $\mathcal{L}_{\rm{SCR}}^{\rm{img}}$ in Eqn.~(\ref{eq:mcr_img}) for cross-domain semantic knowledge sharing. Based on $3^\dag$, adding with $\mathcal{L}_{\rm{SCR}}^{\rm{dataset}}$ or $\mathcal{L}_{\rm{SCR}}^{\rm{img}}$ can improve the mIoU by $0.43\%$ or $0.91\%$ respectively, which shows that two kinds of semantic consistency regularization can both boost performance. Meanwhile, jointly adding the $\mathcal{L}_{\rm{SCR}}^{\rm{dataset}}$ and $\mathcal{L}_{\rm{SCR}}^{\rm{dataset}}$ can further improve the performance to $72.25\%$ mIoU.

\begin{table}
\centering
\resizebox{\linewidth}{!}{
\begin{tabular}{ccccccc} 
\toprule [1 pt]
 \multirow{2}{*}{No.} & \multirow{2}{*}{Multi-datasets} & 
 \multicolumn{2}{c}{MSE} & \multicolumn{2}{c}{MST}  & \multirow{2}{*}{mIoU ($\%$)} \\  \cline{3-6}
  &  & $\mathcal{L}_{\rm{aux}}$ (None) & $\mathcal{L}_{\rm{aux}}$ ($\mathbf{M}_{\rm{intra}}$) & $\mathcal{L}_{\rm{SCR}}^{\rm{dataset}}$ & $\mathcal{L}_{\rm{SCR}}^{\rm{img}}$ & \\ \hline
\textbf{1$^{\dag}$}&$\checkmark$& & & & & 70.40\\
\textbf{2$^{\dag}$}&$\checkmark$&$\checkmark$ & & & &  70.78\\
\textbf{3$^{\dag}$}&$\checkmark$ & &$\checkmark$ & & & 71.02\\
\textbf{4$^{\dag}$}&$\checkmark$ & &$\checkmark$ &$\checkmark$ & &  71.45\\
\textbf{5$^{\dag}$}&$\checkmark$ & &$\checkmark$ &&$\checkmark$  & 71.93\\ 
 \textbf{6$^{\dag}$}&$\checkmark$ & &$\checkmark$ &$\checkmark$ & $\checkmark$ & \textbf{72.25}\\
\bottomrule[1 pt]
\end{tabular}}
\vspace{-0.2cm}
\caption{Ablation studies of SST's key components on the PASCAL dataset with ResNet-50, under the universal parsing setting.}
\label{table:ablation}
\vspace{-0.3cm}
\end{table}

\section{Conclusion}
We propose a novel training paradigm called Scalable Semantic Transfer (SST) that aims to leverage the mutual benefits of data from different label domains to address two common parsing scenarios: universal parsing and dedicated parsing. SST introduces three plug-and-play modules that embed prior knowledge of human body parts into a given parsing network during training, which can be removed during inference to avoid piling up extra reasoning costs. The effectiveness of SST is well-demonstrated through extensive experiments on three human parsing benchmarks.

\section*{Acknowledgement}
The work is supported in part by the Young Scientists Fund of the National Natural Science Foundation of China under grant No.62106154, by the Natural Science Foundation of Guangdong Province, China (General Program) under grant No.2022A1515011524, by CCF-Tencent Open Fund, by Shenzhen Science and Technology Program ZDSYS20211021111415025, and by the Guangdong Provincial Key Laboratory of Big Data Computing, The Chinese University of Hong Kong (Shenzhen).

{\small
\bibliographystyle{ieee_fullname}
\bibliography{egbib}
}
\end{document}